# BreastXAgent: A lightweight agentic explainability framework for breast density classification

Muhammad Turab[a], Raja Vavekanand*[b], and Knooz Fatima[c]
[a]School of Computing, Dublin City University, Dublin, D09 Y07, Ireland
[b]Department of Information Technology, Benazir Bhutto Shaheed University Lyari, Karachi, 75660, Pakistan
[c]Iqra University, Hyderabad, Pakistan
Email: rvavekan@gmail.com

## ABSTRACT

Deep learning models for breast density classification typically return only a label and confidence score. This paper presents BreastXAgent, a lightweight agentic framework that combines a pretrained classifier, Grad-CAM visual explanation, LangGraph-based routing, LLM report generation, and rule-based guardrails to produce controlled, faithful, non-diagnostic reports. On 160 mammography images, uncontrolled LLM reporting produced 75% Grad-CAM overclaims and 62.5% clinical overclaims. Strict prompting and LangGraph routing improved safety but did not eliminate violations. The full BreastXAgent achieved 100% safety pass rate, zero overclaims, and highest utility score. Results demonstrate that agentic guardrails enable safe, structured explanation generation without modifying the underlying classifier.

**Keywords:** Agentic AI, breast density classification, Grad-CAM, explainable AI, LangGraph

## 1. INTRODUCTION

Breast imaging is a vital application where mammographic breast density classification plays an important role due to several reasons: firstly, dense breast tissue can decrease the mammographic sensitivity; secondly, dense breast tissue is linked to higher breast cancer risk [1,2]. Breast density typically appears in clinical reporting with the four BI-RADS categories: almost entirely fatty, scattered fibroglandular density, heterogeneously dense, and extremely dense [3]. These categories have been with good potential to be automatically predicted from mammogram images using deep learning models [4,5]. But, many classification systems still provide only the predicted class and a level of confidence, which are not very helpful for providing understandable, confidence aware explanations of model behavior.

Explainable AI methods overcome this aspect by generating visual evidence. Grad-CAM is a gradient-based method for creating heatmaps on the convolutional neural network for a specific class in the image [6]. These heatmaps can be used in mammography to check the regions of the image used to predict the density. While useful, it is important to note that Grad-CAM does not offer structured reasoning and should not be used as anatomical proof, as a tissue localisation, or as a diagnostic evidence. Interpretation is still required to be cautious.

Another recent path has been to medical AI agents: agentic systems orchestrate multiple tools, decisions and validation steps. In medical imaging reasoning, medRAX formalises the medical imaging reasoning problem as a tool-orchestration problem [7]. LangGraph provides these workflows by providing shared state, nodes and edges where each node computes some routing and logic to decide what to do next [8]. Such an architecture is appropriate for an explanation pipeline where prediction, evidence handling, report generation and safety validation are separated.

In this paper, we introduce a lightweight proof-of-concept framework for controlled breast-density explanation, BreastXAgent. The system integrates a pre-trained MONAI breast-density classifier [9], a Grad-CAM visual explanation, a LangGraph system for routing confidence and evidence, and rule-based safeguards to prevent medical or visual claims that are not supported by the system. The aim is not to propose a new classifier or diagnostic system, but to explore if an agentic workflow can convert model outputs and Grad-CAM artifacts into structured, faithful, confidence-aware, non-diagnostic research reports. In contrast to previous approaches that rely on direct text generation in medical LLM reporting, BreastXAgent proposes three novel contributions: (1) Confidence-conditioned agentic routing, (2) Guardrail replacement (not warning) of unsafe outputs, and (3) Modular separation of attribution, confidence, and language generation without the need to retrain a classifier. This paper makes the following main contributions:

(a) A modular agentic explanation pipeline for breast-density classification combining prediction, visual attribution, workflow routing, LLM generation, and guardrail validation.

(b) BreastReportBench, a report-generation evaluation setting comparing template-only, LLM-only, strictprompt, LangGraph-based, and full guardrailed strategies.

(c) Demonstration that uncontrolled LLM reporting produces unsupported Grad-CAM and clinical overclaims, while guardrail validation and deterministic fallback improve final report safety and faithfulness.

## 2. RELATED WORK

Breast density classification is commonly formulated using the four BI-RADS density categories: A, B, C, and D. Recent deep learning approaches typically treat this as an image-level classification task, where a CNN-based model predicts the density category directly from the mammogram. In this work, we use the pretrained MONAI breast-density classification model as the visual inference component [9]. The model provides a practical baseline because it outputs the four density classes without task-specific retraining. However, a standalone classifier only provides a predicted label and confidence score, without explaining the visual evidence behind the decision or converting the result into a controlled clinical-style report.

Visual explanation methods are often used to make CNN decisions more interpretable by highlighting image regions that contribute to a predicted class [10]. Grad-CAM is one of the most widely used methods for this purpose, as it uses class-specific gradients to weight convolutional feature maps and generate a coarse localisation heatmap [6]. In our system, Grad-CAM is applied to the final convolutional block of the Inception-v3 backbone and overlaid on the mammogram to show the regions influencing the density prediction. While Grad-CAM can provide useful visual evidence, it does not by itself produce structured reasoning, confidence interpretation, or validated textual reporting. Therefore, it is used here as an evidence source within a larger agentic reporting workflow.

Agentic AI systems extend beyond single-model prediction by coordinating multiple modules through a controlled workflow. LangGraph supports this type of design by representing the workflow as a state-based graph, where each node updates and passes shared state information [8,11]. In our work, this structure connects density prediction, confidence analysis, Grad-CAM evidence, LLM-based report generation, and report validation. The reporting node uses Meta-Llama-3.1-8B-Instruct, an instruction-tuned text model designed for text-in/text-out dialogue tasks [12]. MedRAX is a relevant reference point because it presents medical-imaging AI as a toolorchestration problem rather than only a classification problem [7]. Inspired by this direction, BreastXAgent provides a compact proof-of-concept for mammography reporting. Unlike a standalone classifier or heatmap generator, it combines prediction, visual explanation, language generation, and guardrail validation to produce controlled breast-density explanation reports.

## 3. METHODOLOGY

BreastXAgent is a framework that is based on modules of an agentic explanation layer for breast density classification. The system consists of five steps as illustrated in figure 1, visual prediction, visual explanation, workflow routing, language generation, and guardrail validation. Every step runs on the same state object, which is orchestrated by LangGraph, allowing to implement deterministic steps and fallback conditions.

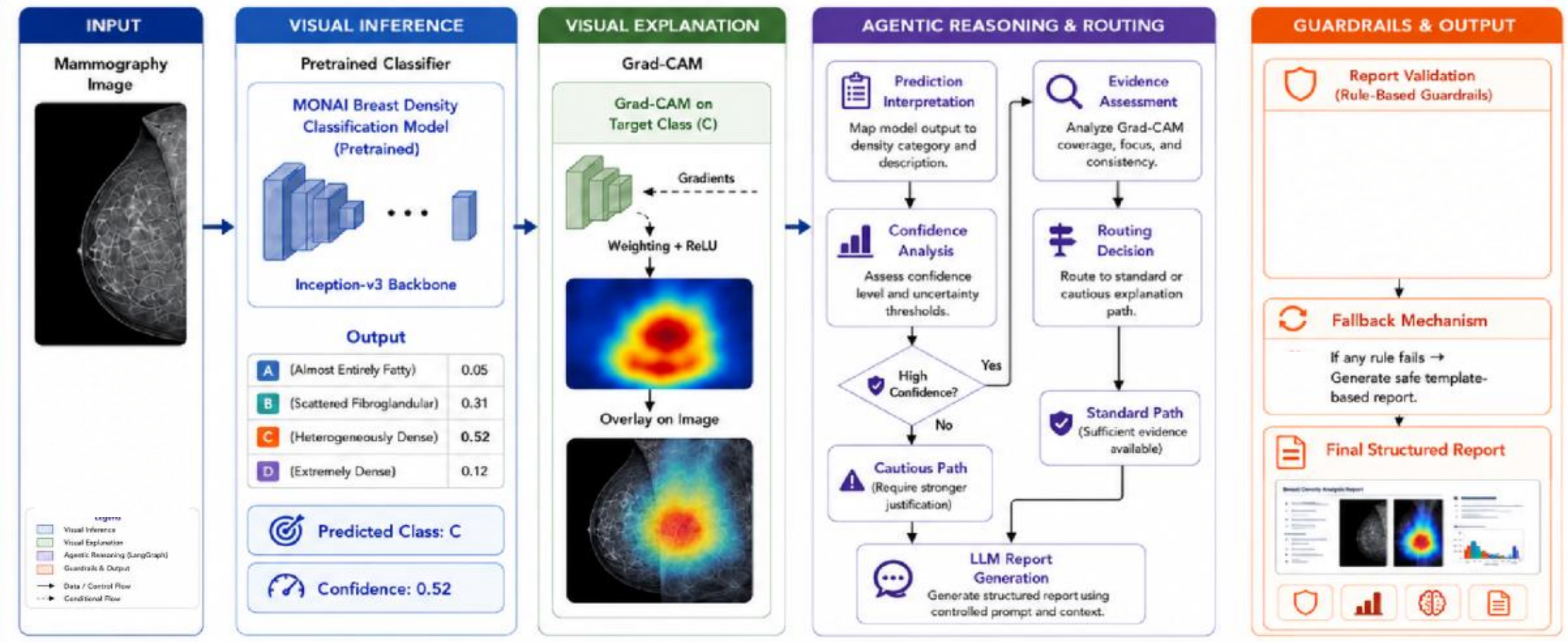

Figure 1. Architectural overview of the BreastXAgent explanation pipeline. The classifier yields a density label and confidence. Grad-CAM generates a heatmap. LangGraph routes based on confidence level. LLM generates a first draft report. The guardrail ensures safety and provides deterministic fallback if necessary.

### 3.1 Classification and visual explanation

The visual inference module uses a pretrained breast density classifier from MONAI (Hugging Face repository [9]). The model accepts a mammogram image $\mathbf{X} \in \mathbb{R}^{H\times W}$ and outputs a probability vector $\mathbf{p} = [p_A, p_B, p_C, p_D]$ over the four BI-RADS density classes: A (almost entirely fatty), B (scattered fibroglandular), C (heterogeneously dense), D (extremely dense). The predicted class is:

$$\hat{y} = \arg \max_{k\in\{A,B,C,D\}} p_k. \tag{1}$$

The confidence score is defined as $c = \max(\mathbf{p})$.

Grad-CAM is applied to the final convolutional layer of the Inception-v3 backbone. For the predicted class $\hat{y}$, the gradient of the class score $S_{\hat{y}}$ with respect to feature map activations $A^l_{ij}$ (at layer $l$, spatial location $(i,j)$) is computed. The importance weight $\alpha_k^l$ for feature map $k$ is:

$$\alpha_k^l = \frac{1}{Z}\sum_i\sum_j \frac{\partial S_{\hat{y}}}{\partial A^l_{ij,k}}, \tag{2}$$

where $Z$ is the number of spatial locations. The Grad-CAM heatmap $H$ is:

$$H = \mathrm{ReLU}\left(\sum_k \alpha_k^l A_k^l\right). \tag{3}$$

The heatmap $H$ is resized to $H \times W$ and overlaid on $\mathbf{X}$ to produce a visual explanation artifact.

### 3.2 Confidence categorisation and routing

The raw confidence score $c$ is mapped to a discrete confidence level $\ell$ using fixed thresholds:

$$\ell = \begin{cases} \text{high}, & c \geq 0.85 \\ \text{moderate}, & 0.70 \leq c < 0.85 \\ \text{low}, & c < 0.70 \end{cases} \tag{4}$$

The LangGraph router selects a report generation path based on $\ell$:

**High confidence**: Full report with specific density description.

**Moderate confidence**: Report includes a qualification phrase (e.g., "moderate confidence").

**Low confidence**: Report includes a disclaimer and avoids detailed visual interpretation.

Algorithm 1 formalises the routing logic.

**Algorithm 1** LangGraph Routing Decision

```
Input: Predicted class ŷ, confidence c, Grad-CAM file G
Compute ℓ from Eq. (4)
Initialize shared state S ←{y,c,ℓ,Ĝ}
if ℓ == high then
        Set routing ← FullReport
else if ℓ == moderate then
        Set routing ← QualifiedReport
else
        Set routing ← DisclaimerReport
end if
Append routing decision to S
Pass S to LLM generation node
```

### 3.3 LLM report generation

The language generation node is based on the Meta-Llama-3.1-8B-Instruct model [12] with a prompt for strict instructions. The LLM takes an input that is a structured prompt with. Predicted class $\hat{y}$ and description of density. Confidence score c and level $\ell$. As a reference (not the image), write the name of the file you are using in the graduate school. Evidence status: "Relative activation", not tissue composition: Evoked by Evidenced by: Evidenced by: ("Grad-CAM indicates relative activation, not tissue composition.") Agent decision (from algorithm 1). The LLM is clearly directed to:

(a) Don't display a visual interpretation of the mammogram or the Grad-CAM.

(b) Do not make any claims to diagnostic or anatomical localisation.

(c) Vary them according to the confidence level (e.g. "likely" for moderate, "may" for low).

The report contains 3 parts: **Findings, Visual Evidence, and Impression.**

### 3.4 Guardrail validation and fallback

Once the text has been generated using LLM, a rule-based guardrail module validates the text. Unsupported claims are identified via keyword- and pattern-matching:

(a) Diagnosis: "cancer", "malignant", "benign".

(b) Tissue localisation: "in the upper upper quadrant", "located at".

(c) Grad-CAM overclaim: "proves", "shows tissue composition", "clinical evidence".

(d) Recommendations: "follow-up", "biopsy", "additional imaging".

If there is any pattern that is not allowed, the output of the LLM is discarded. A deterministic fall-back report is produced in its place.

The model is [low/moderate/high] confident of density class [X] (with score [c]). The GradCAM was generated but not interpreted here. This is a research output and not a clinical diagnosis."

Let $R_{\text{LLM}}$ be the generated report and $R_{\text{fallback}}$ be the deterministic template. The final report $R_{\text{final}}$ is:

$$R_{\text{final}} = \begin{cases} R_{\text{LLM}}, & \text{if guardrail}(R_{\text{LLM}}) = \text{pass} \\ R_{\text{fallback}}, & \text{otherwise} \end{cases} \quad (5)$$

Algorithm 2 summarises the validation process.

**Algorithm 2** Guardrail Validation and Fallback

```
Input: LLM output R_LLM, forbidden pattern set F
R_LLM ← lowercase(R_LLM)
violation ← False
for each pattern f ∈ F do
        if f in R_LLM then
        violation ← True
        break
        end if
end for
if violation is True then
        R_final ← generateFallback()
else
R_final ← R_LLM
end if
return R_final
```

# 4. EXPERIMENTS AND RESULTS

The experiment used 160 mammography images from a private sample set (class distribution: 12.5% A, 43.75% B, 37.5% C, 6.25% D). Five reporting strategies were used to process each image:

**(a) Template-only:** Report generation without LLM (deterministic).

**(b) LLM-only:** directly generating the LLM.

**(c) Strict-prompt LLM:** LLM generation that is prompted by strict safety prompts.

**(d) LangGraph + LLM:** Routed report generation without fallback replacement.

**(e) BreastXAgent:** Full workflow along with routing, guardrail and deterministic fallback.

For the LLM inferences, Meta-Llama-3.1-8B-Instruct was used with temperature = 0.0 to ensure deterministic results. The metrics used for evaluation are given in table 1.

Table 1. Evaluation Metrics for Report Generation Strategies.

| Metric | Definition |
|---|---|
| Completion rate | Reports containing all three required sections |
| Input faithfulness | Preserves predicted class, confidence score, density description |
| Confidence consistency | Report wording consistent with assigned confidence level |
| Grad-CAM overclaim rate | Reports overstating Grad-CAM as proof or clinical evidence |
| Clinical overclaim rate | Reports containing unsupported diagnostic claims |
| Safety pass rate | Reports passing all faithfulness, confidence, and non-diagnostic checks |
| Fallback rate | Reports replaced by deterministic fallback (BreastXAgent only) |
| Utility score | Composite score (0–100%) combining completeness, safety, confidence awareness |

### 4.1 Reporting strategy comparison

The benchmark consists of 160 images for which five reporting strategies are compared in table 2. The LLM-only baseline shows mediocre performance, while 12.5% of them pass the safety test and 75% of them have an overclaim in Grad-CAM. The use of strict prompting brings the number of overclaims down to 12.5% but still not enough. With LangGraph, the confidence consistency is raised to 93.8%, and with BreastXAgent, the safety and utility are raised to 100% with the use of a guardrail mechanism to fall back to deterministic safe reports.

Table 2. Reporting Strategy Comparison on BreastReportBench (N=160).

| Method | Completion | Faithfulness | Conf. Consistency | Grad-CAM Overclaim | Clinical Overclaim | Safety Pass | Utility |
|---|---|---|---|---|---|---|---|
| Template-only | 100.0% | 100.0% | 100.0% | 0.0% | 0.0% | 100.0% | 68.8% |
| LLM-only | 68.8% | 81.3% | 56.3% | 75.0% | 62.5% | 12.5% | 31.3% |
| Strict-prompt LLM | 100.0% | 100.0% | 87.5% | 12.5% | 6.3% | 81.3% | 81.3% |
| LangGraph + LLM | 100.0% | 100.0% | 93.8% | 6.3% | 6.3% | 87.5% | 87.5% |
| BreastXAgent | 100.0% | 100.0% | 100.0% | 0.0% | 0.0% | 100.0% | 100.0% |

### 4.2 Ablation study

To assess each module's contribution, we performed an ablation study by removing components from BreastXAgent. As shown in table 3, removing the strict prompt caused complete safety failure (0%). Without the guardrail, safety dropped to 87.5% due to unreplaced unsafe drafts. Removing LangGraph routing preserved safety (93.8%) but reduced utility to 75%, while template-only reporting was safe but had the lowest utility (68.8%) due to limited adaptive explanation.

Table 3. Ablation Study for Report Generation Strategies (N=160).

| Variant | Removed Component | Safety Pass ↑ | Grad-CAM Overclaim ↓ | Utility ↑ |
|---|---|---|---|---|
| Full BreastXAgent | None | 100.0% | 0.0% | 100.0% |
| Without guardrail | Guardrail fallback | 87.5% | 6.3% | 87.5% |
| Without LangGraph routing | LangGraph routing | 93.8% | 6.3% | 75.0% |
| Without strict prompt | Strict prompt | 0.0% | 100.0% | 25.0% |
| Template-only | LLM generation | 100.0% | 0.0% | 68.8% |

### 4.3 Guardrail intervention analysis

Grad-CAM overclaims were most common (4.4%), followed by tissue localisation claims (3.1%). No diagnostic or recommendation violations occurred. The 6.3% intervention rate shows that most reports (93.7%) passed unchanged. Table 4 shows the distribution of guardrail intervention triggers across the 160 cases in the BreastXAgent configuration.

Table 4. Guardrail violation trigger rates in BreastXAgent (N=160). The guardrail replaced 6.3% of LLM-generated reports with deterministic fallback.

| Violation Type | Trigger Rate |
|---|---|
| Diagnosis terms (cancer, malignant) | 0.0% |
| Tissue localization claims | 3.1% |
| Grad-CAM overclaims | 4.4% |
| Recommendation statements | 1.9% |
| Any violation | 6.3% |

### 4.4 Qualitative example

LLM-only outputs show localisation, clinical interpretation, and recommendation violations. Strict prompting reduces overclaims but remains insufficiently detailed. BreastXAgent provides the most comprehensive, secure, confidence-aware, and clearly non-diagnostic report. Table 5 compares reports for a class B case with moderate confidence (0.78), while figure 2 gives a clearer visual explanation.

Table 5. Qualitative Report Comparison for a Single Case (Class B, Confidence 0.78)

| **Method** | **Generated Report (Impression Section)** |
|---|---|
| LLM-only | "Grad-CAM clearly shows scattered fibroglandular density localised to the upper outer quadrant. This finding is suggestive of benign breast changes. Follow-up may be considered." |
| Strict-prompt LLM | "The model predicts scattered fibroglandular density with moderate confidence (0.78). Grad-CAM indicates relative activation, not tissue proof." |
| BreastXAgent | "The model predicts class B (scattered fibroglandular density) with moderate confidence (0.78). Grad-CAM was generated as a visual attribution artifact but is not interpreted as tissue localisation. This is a research report and not a clinical diagnosis." |

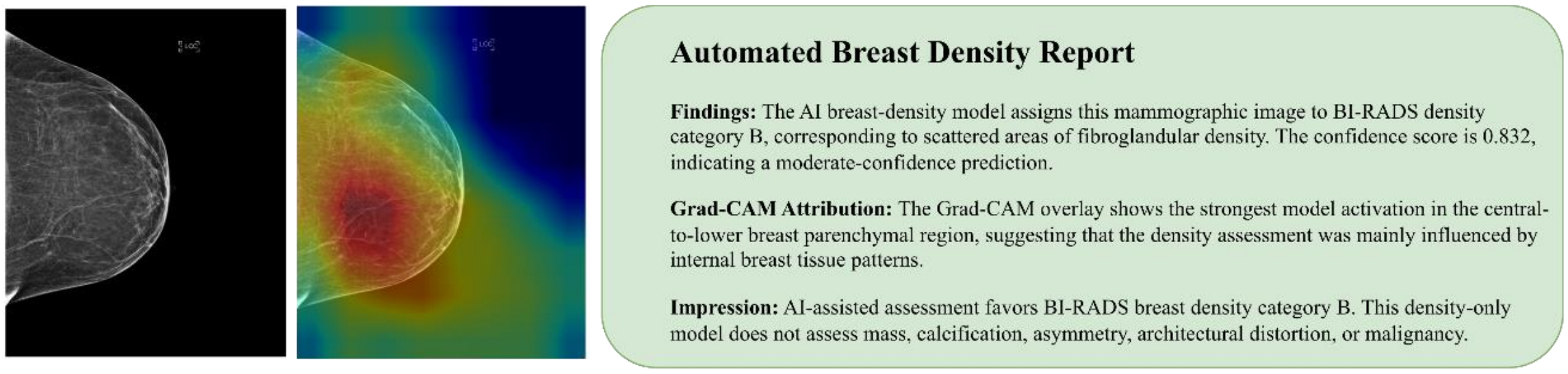

**Input** **GradCAM** **BreastXAgent Report**

Figure 2. BreastXAgent framework output with safety guardrails. From left to right: the input mammogram, the GradCAM attribution map, and the generated report. The output demonstrates accurate BI-RADS density prediction while effectively suppressing hallucinated or unsupported diagnostic claims through strict guardrailing.

## 5. DISCUSSION

The findings reveal that the reporting of uncontrolled LLM is unsafe 75% of reports is over-claimed with the use of Grad-CAM, and in 62.5% of the cases the clinical style outputs. Whilst strict prompting did decrease, but not abolish these violations, this confirmed that prompts are not sufficient on their own for safety critical reporting. Confidence consistency was enhanced by LangGraph routing, and 100% safety was realised by guardrails and deterministic fallback in the entire BreastXAgent pipeline, with 6.3% of reports that needed to be intervened. Ablation results indicate that without the use of strict prompt, it was necessary for 0% of the patients to pass the safety section, whereas the use of the template alone resulted in the highest safety score (68.8%) and the least helpful (lowest utility score). However, the dataset of 160 images is small, a single classifier is used and there are rule-based guardrails that can be missing semantic overclaims. Larger datasets, incorporation of classifier uncertainty in routing and the design of repair based guardrails are some of the future works that should be considered.

## 6. CONCLUSION

This paper introduced BreastXAgent, a lightweight agentic explainability framework for breast density classification that combines a pretrained MONAI classifier, Grad-CAM, LangGraph routing, LLM report generation, and rule-based guardrails. Using 160 mammography images, the full pipeline achieved 100% safety and utility, with zero clinical or Grad-CAM overclaims, while LLM-only reporting passed safety checks in only 12.5% of cases. The guardrail fallback corrected unsafe wording in 6.3% of reports, showing that BreastXAgent's main contribution is controlled, faithful, and non-diagnostic explanation generation rather than improved classifier accuracy. Future work will evaluate larger datasets, add classifier error handling, and develop repair-based guardrails to reduce fallback dependence.

## REFERENCES

[1] Edmonds, C. E., O'Brien, S. R., and Conant, E. F., "Mammographic breast density: Current assessment methods, clinical implications, and future directions," Seminars in Ultrasound, CT and MRI 44(1), 35–45 (2023).
[2] Turab, M. and Jamil, S., "A comprehensive survey of digital twins in healthcare in the era of metaverse," BioMedInformatics 3(3), 563–584 (2023).
[3] Acciavatti, R. J., Lee, S. H., Reig, B., Moy, L., Conant, E. F., Kontos, D., and Moon, W. K., "Beyond breast density: Risk measures for breast cancer in multiple imaging modalities," Radiology 306(3), e222575 (2023).
[4] Abhisheka, B., Biswas, S. K., and Purkayastha, B., "A comprehensive review on breast cancer detection, classification and segmentation using deep learning," Archives of Computational Methods in Engineering 30, 5023–5052 (2023).
[5] Chakravarthy, S. S. R. and Rajaguru, H., "Automatic detection and classification of mammograms using improved extreme learning machine with deep learning," IRBM 43(1), 49–61 (2022).
[6] Selvaraju, R. R., Cogswell, M., Das, A., Vedantam, R., Parikh, D., and Batra, D., "Grad-cam: Visual explanations from deep networks via gradient-based localization," in [Proceedings of the IEEE International Conference on Computer Vision], 618–626 (2017).
[7] Fallahpour, A., Ma, J., Munim, A., Lyu, H., and Wang, B., "Medrax: Medical reasoning agent for chest x-ray," arXiv preprint arXiv:2502.02673 (2025).
[8] LangChain, "Langgraph graph api overview." https://docs.langchain.com/oss/python/langgraph/ graph-api.
[9] MONAI, "Breast density classification." Hugging Face model repository. https://huggingface.co/MONAI/ breast_density_classification/tree/0.1.7.
[10] Yeh, W.-C., Shia, W.-C., Hsu, Y.-T., Huang, C.-H., and Lee, Y.-S., "A lightweight breast cancer mass classification model utilizing simplified swarm optimization and knowledge distillation," Bioengineering 12(6), 640 (2025).
[11] Arunkumar, V., Gangadharan, G. R., and Buyya, R., "Agentic artificial intelligence (ai): Architectures, taxonomies, and evaluation of large language model agents," arXiv preprint arXiv:2601.12560 (2026).
[12] Meta, "Llama-3.1-8b-instruct." Hugging Face model card. https://huggingface.co/meta-llama/ Llama-3.1-8B-Instruct.